\documentclass{article}

\usepackage[preprint]{neurips_2024}

\usepackage[utf8]{inputenc} 
\usepackage[T1]{fontenc}    
\usepackage{hyperref}       
\usepackage{url}            
\usepackage{booktabs}       
\usepackage{amsfonts}       
\usepackage{nicefrac}       
\usepackage{microtype}      
\usepackage{xcolor}         
\usepackage{amsmath}        
\usepackage{graphicx}       

\title{Convergence of Multiagent Learning Systems for Traffic control}

\author{%
  Sayambhu Sen \\
  Amazon Alexa\\
  \texttt{sensayam@amazon.com} \\
  \And
  Shalabh Bhatnagar\\
  Indian Institute of Science \\
  \texttt{shalabh@iisc.ac.in} \\
}

\begin{document}

\maketitle

\begin{abstract}
  Rapid urbanization in cities like Bangalore has led to severe traffic congestion, making efficient Traffic Signal Control (TSC) essential. Multi-Agent Reinforcement Learning (MARL), often modeling each traffic signal as an independent agent using Q-learning, has emerged as a promising strategy to reduce average commuter delays. While prior work \cite{prashanttraffic} has empirically demonstrated the effectiveness of this approach, a rigorous theoretical analysis of its stability and convergence properties in the context of traffic control has not been explored. This paper bridges that gap by focusing squarely on the theoretical basis of this multi-agent algorithm. We investigate the convergence problem inherent in using independent learners for the cooperative TSC task. Utilizing stochastic approximation methods, we formally analyze the learning dynamics. The primary contribution of this work is the proof that the specific multi-agent reinforcement learning algorithm for traffic control is proven to converge under the given conditions extending it from single agent convergence proofs for asynchronous value iteration. 

\end{abstract}

\section{Introduction}
Real-time traffic evolves according to a complex stochastic process. Following \cite{prashanttraffic}, we frame the problem of optimizing green signal durations as a minimization of overall road occupancy, which we assume will be the strategy for maximizing traffic flow. While this problem can be framed as a single Markov Decision Process (MDP), a centralized, single-agent approach—where the state space represents all cars in the network—is intractable. As the number of junctions increases, this centralized state space explodes exponentially.

We therefore adopt a Multi-Agent  Reinforcement Learning (MARL) approach, which reduces the state space to a manageable size for each agent. Each agent then only needs to explore its local state space. This strategy not only reduces computational overhead but also facilitates faster convergence through more focused exploration.

We model the Traffic Signal Control (TSC) problem as a multi-agent coordination challenge. While previous simulation studies have demonstrated the efficacy of MARL for optimal signal timing, this work focuses on the theoretical analysis. Our primary objective is to model the convergence of the MARL Q-learning algorithm and derive the formal conditions under which it converges.

The paper is organized as follows:
\begin{itemize}
    \item First, we present the Q-learning algorithm as applied in simulation studies.
    \item Second, we develop a theoretical model using a simple three-junction network, applying stochastic modeling techniques to formulate preliminary convergence conditions.
    \item Finally, we introduce the stochastic approximation method, modeling our system as a discrete Euler approximation of an ordinary differential equation (ODE). We then show the conditions for the ODE's convergence and discuss the implications for both single-agent and multi-agent systems.
\end{itemize}

\section{Modelling and Convergence}
\subsection{Modelling the MDP and the Q-Learning algorithm}
We will describe the discrete state and action spaces of the MDP to formulate the Q-learning algorithm. For the given traffic network, we will model each junction's state space and action space separately. Each junction is assumed to be connected to neighbouring junctions via 2-way roads. A state $s_j$ for a given junction $j$ is defined as a vector of dimension $L+1$, where $L$ denotes the number of incoming lanes to that junction. The $i$-th component of the state vector, $q^j_i$ (where $i \in \{1, 2, \ldots, L\}$), denotes the queue length of the traffic in the $i$-th lane of that junction. The final component, $q^j_{L+1}$, denotes the index of the phase (or road) that has been set to green for incoming traffic in the round-robin (RR) schedule of the traffic controller.


This state vector can be explicitly written as:
$$ \mathbf{s}_j = (q^j_1, q^j_2, \ldots, q^j_L, q^j_{L+1}) $$

The effective state space for the entire network is the Cartesian product of the individual state spaces for each junction. If $S_j$ denotes the set of all states for junction $j$, then the complete state space is given by $S = \times_{j=1}^{N} S_j$. If a single, centralized agent were to map actions for every state in this composite space, the complexity of the problem would rise rapidly.

To prevent this, we utilize a decentralized approach with separate agents, where each agent explores only its own junction's state space. However, an agent's actions still affect its local neighbors; thus, the system's dynamics indirectly incorporate the joint actions of neighboring agents.

Since this is equivalent to each agent observing only a part of the state space, we can model this as a Partially Observable Markov Decision Process (POMDP). However, despite considerably reducing the state space size using a decentralized setting, the cardinality of the state space (the number of possible states) is still upper-bounded by $q_{max}^{L}$. This value, representing the maximum number of vehicles a lane can accommodate, can be exceedingly large for longer roads.

To address this, we segment the road based on occupancy rather than using the exact vehicle count. We discretize the queue length into three segments: \{low = 0, medium = 1, high = 2\}. This characterization is achieved by choosing two distance thresholds, $D_1$ and $D_2$, from the traffic junction. The traffic level is then classified as \{low, medium, high\} based on whether the queue length has reached these threshold distances. The traffic is characterized as follows:
\begin{equation}
    q_i^j(t) = 
    \begin{cases} 
          0, & \text{if } q_i^{j'} < D_1 \\
          1, & \text{if } D_1 \le q_i^{j'} < D_2 \\
          2, & \text{if } D_2 \le q_i^{j'}
    \end{cases}
    \label{eq:occupancy}
\end{equation}

where $q^{j'}_i$ denotes the actual congestion level at time $t$ for junction $j$ on lane $i$.

Thus, the state vector $\mathbf{s}_j(t)$ for a given junction $j \in \{1, 2, \ldots, J\}$ at time $t$ is a vector of the discretized queue lengths for each incoming lane of that junction. It is denoted by:
$$ \mathbf{s}_j(t) = (q^j_1(t), \ldots, q^j_{L_j}(t), q^j_{L_j+1}(t))^T $$
Here, the last component, $q^j_{L_j+1}(t)$, denotes the currently active phase in the round-robin (RR) schedule. It is to be noted that outgoing lanes are not explicitly included in this state vector. This is acceptable because each outgoing lane from one junction is considered an incoming lane for another junction, ensuring all roads in the network are covered.

In a similar manner, we also discretize the action space. The action for an agent $j$ is the duration of the green phase in the RR schedule. We do not need to model the *sequence* of the RR schedule, as the sequence is considered fixed and is already accounted for in the state vector (in the $q^j_{L_j+1}(t)$ component).

Thus, while the action space could have been a continuous regression output representing the time a phase is green, we reduce its size in a similar spirit to the state space reduction. We discretize the action $a_j(t)$ into one of three values: \{low, medium, high\}. For simulation studies, these values are typically set to: low = 10 seconds, medium = 20 seconds, and high = 30 seconds.

We are considering a stationary policy for our methods, such that $\pi_j = \{\mu_j, \mu_j, \ldots\}$, where $\mu_j$ is a mapping from the state $\mathbf{s}_j(t)$ to one of the discrete actions. Thus, $\mu_j(\mathbf{s}_j(t))$ is the mapping that sets the time duration of the active phase at junction $j$ given the state $\mathbf{s}_j(t)$. All actions are assumed to be feasible at every state. The setting is designed such that the policies obtained for each junction satisfy the goal of minimizing the long-term average delay for all vehicle users.

Finally, the cost signal $c_j(t)$ for an agent at junction $j$ is determined. This cost is observed after taking action $\mu_j(\mathbf{s}_j(t))$ at state $\mathbf{s}_j(t)$, which leads to the next state $\mathbf{s}_j(t+1)$. It is defined as the average occupancy of all lanes $i$ across all junctions $k$ in the agent's neighborhood $N_j$ (which typically includes $j$ itself):
\begin{equation}
    c_j(t) = \frac{1}{|N_j|} \sum_{k \in N_j} \sum_{i=1}^{L_k} q_i^k(t+1)
    \label{eq:cost}
\end{equation}
Here, $q_i^k(t+1)$ represents the discretized occupancy of lane $i$ at junction $k$ at the next time step, as defined in \eqref{eq:occupancy}. $N_j$ denotes the set of neighboring junctions to junction $j$, as well as junction $j$ itself. Consequently, $|N_j|$ denotes the cardinality (or the number of elements) in the set $N_j$. This cost function is designed in such a way that each agent is forced to take into account the effect of its actions on the neighboring junctions. The cost $c_j(t)$ measures how the queue lengths of neighboring junctions are affected at time $t+1$ by choosing action $a_j(t)$ at time $t$.

Each agent, corresponding to its junction, thus obtains cost feedback signals from its neighbors and uses them to update the Q-values for the Q-learning algorithm. The exploration strategies used for the agents are $\epsilon$-greedy and the UCB algorithm. The next section describes the Q-learning algorithm used for learning the state values at each junction, as well as the exploration strategies employed.

\subsection{Q-Learning algorithm}
Q-values are assigned to each state-action pair $(s, a)$. It is these Q-values that determine the optimal policy for the TSC problem. The Q-factor for a policy $\pi$, given by $Q^{\pi}(s, a)$, is a measure of how "good" it is to take action $a$ in state $s$ and subsequently follow policy $\pi$.

When the Q-function is optimal, $Q^{*}(s, a)$ gives the minimum expected sum of discounted single-stage costs achieved by choosing action $a$ in state $s$ at time 0 and subsequently following the optimal policy $\pi^{*} = \{\mu^{*}, \mu^{*}, \ldots\}$ thereafter. This is expressed as:
\begin{equation}
    Q^{*}(s, a) = k(s, a) + E\left[ \sum_{t=1}^{\infty} \beta^t k(s_t, \mu^{*}(s_t)) \, \middle| \, s_0 = s, a_0 = a, \mu^{*} \right], \quad \forall (s, a)
    \label{eq:q_optimal}
\end{equation}
where $k(s, a)$ is the immediate or single-stage cost for taking action $a$ in state $s$, and $\beta$ is the discount factor.

In order to find the optimal policy, we must first find the optimal Q-function, $Q^{*}(s, a)$, for all state-action pairs. Once the optimal Q-function is obtained, the policy that minimizes the long-term discounted cost is given by:
\begin{equation}
    \mu^{*}(s) = \arg\min_{a \in A} Q^{*}(s, a)
    \label{eq:optimal_policy}
\end{equation}

In order to find the optimal Q-function, we use an iterative update rule so that the Q-values converge to their optimal values. 

\begin{itemize}
    \item The Q learning algorithm for the single agent system.
    \begin{equation}
        Q_{t+1}(s,a) = Q_t(s,a) + \gamma(t)\left(c(t) + \beta \min_{b \in A} Q_t(s', b) - Q_t(s,a)\right) 
        \label{eq:equationQsingleagent}
    \end{equation}

    \item The Q Learning algorithm for the Multi-Agent system. \cite{prashanttraffic}
    \begin{equation}
        Q^j_{t+1}(s^j, a^j) = Q^j_t(s^j, a^j) + \gamma(t)\left(c_j(t) + \beta \min_{b \in A} Q^j_t(s^{j'}, b) - Q^j_t(s^j, a^j)\right) 
        \label{eq:equationQmultiagent}
    \end{equation}
\end{itemize}

The agent at junction $j$ follows the update rule:
Let $Q^j_0(s_j, a_j)$ be initialized for all $s_j \in S_j, a_j \in A$. For all $t \ge 0$:
\begin{equation}
\begin{split}
    Q^j_{t+1}(s_j, a_j) = Q^j_t(s_j, a_j) + \gamma(t) \Big( & c_j(t) + \beta \min_{b \in A} Q^j_t(s_j', b) \\
    & - Q^j_t(s_j, a_j) \Big)
\end{split}
\label{eq:q_update}
\end{equation}
where $s_j$ and $s_j'$ refer to the segregated queue length vectors at junction $j$ at times $t$ and $t+1$, respectively, after taking action $a_j$ at time $t$. The step sizes $\gamma(t)$, $t \ge 0$, must satisfy the following standard stochastic approximation conditions:
\begin{equation}
    \gamma(t) > 0, \quad \sum_{t=0}^{\infty} \gamma(t) = \infty, \quad \sum_{t=0}^{\infty} \gamma^2(t) < \infty
    \label{eq:step_size_conditions}
\end{equation}

The exploration strategies used are the $\epsilon$-greedy and Upper Confidence Bound (UCB) algorithms. The $\epsilon$-greedy algorithm, in a given state $s$, chooses the action $a = \arg\min_{a \in A} Q(s, a)$ with probability $1 - \epsilon$, and chooses a random action (uniformly) from $A$ with probability $\epsilon$. One may keep $\epsilon$ fixed, or gradually decay it from an initial value of 1 towards 0, often using an exponential decay rate.

For the UCB exploration strategy, an action is chosen to balance exploitation and exploration. Since this is a cost-minimization problem, this is equivalent to finding the action that maximizes the negative Q-value (exploitation) plus an exploration bonus. The exploration bonus term is inversely proportional to the number of times that action has been chosen in that state. The action is selected by:
\begin{equation}
    a = \arg\max_{c \in A} \left[ -Q^j_t(s_j, c) + \frac{\ln R_{s_j}(t)}{R_{s_j, c}(t)} \right]
    \label{eq:ucb}
\end{equation}
where $R_{s_j}(t)$ denotes the number of times state $s_j$ has been visited up to time $t$, and $R_{s_j, c}(t)$ denotes the number of times action $c$ has been chosen in state $s_j$ up to time $t$.

These two exploration strategies help ensure that all state-action pairs are visited and updated infinitely often, which is a necessary condition for convergence.

\subsection{Example Simple Stochastic Modelling of a 3 junction traffic network}
\label{example3junction}

\begin{figure}[htbp]
    \centering
    \includegraphics[width=0.8\textwidth]{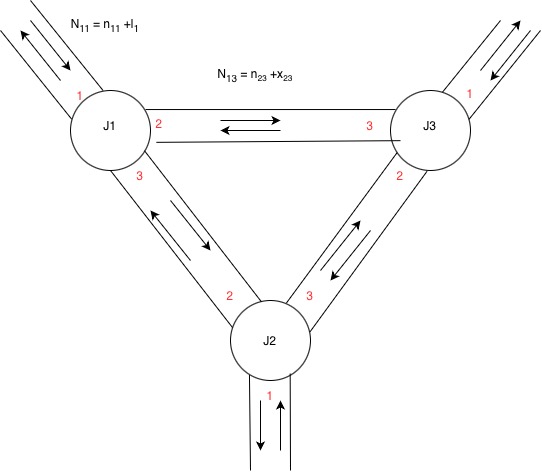}
    \caption{A schematic example of a 3 junction traffic network each connected to each other. The junctions are denoted as $Ji$ and the road attached to each junction is written in \textcolor{red}{red}. Each road is a 2 way road.}
    \label{fig:network_junction}
\end{figure}
We begin with a basic three-junction network where all junctions are interconnected. There are three distinct entry points to the network, one at each junction. Cars are assumed to enter the network according to a Poisson distribution.

It should be noted that for each junction, the incoming lanes are numbered relative to the junction itself in a clockwise manner. At any time $t$, we assume the following equations hold:

For lanes fed by external sources:
\begin{equation}
    N^i_j = n^i_j + I_j
    \label{eq:external_lane}
\end{equation}
where $N^i_j$ represents the current number of cars in the $j$-th junction’s $i$-th lane. The term $n^i_j$ represents the number of cars in that same lane just before the last phase change at junction $j$. The term $I_j$ represents the number of cars arriving from the outside (the external source) at a Poisson rate towards junction $j$. In this particular diagram, the lane numbered '1' at each junction is the external-facing lane from which cars enter the network. Thus, $N^1_j$ is the lane inwards to the junction from the outside.

For lanes fed by other junctions (internal lanes):
\begin{equation}
    N^i_j = n^i_j + x^l_k
    \label{eq:internal_lane}
\end{equation}
where $N^i_j$ and $n^i_j$ have the same meaning as before. However, $x^l_k$ refers to the number of cars in the $k$-th junction’s $l$-th outgoing lane (which feeds lane $i$ of junction $j$).

For each external source $I_j$, cars arrive at an average rate $r_j$. We assume the number of cars $k$ arriving in a time interval $\Delta t$ follows a Poisson distribution with parameter $\lambda = r_j \Delta t$.
\begin{equation}
    P(I_j(\Delta t) = k) = \frac{(r_j \Delta t)^k e^{-r_j \Delta t}}{k!}
    \label{eq:poisson}
\end{equation}
All roads are assumed to have a finite capacity, $K^i_j$:
\begin{equation}
    0 \le N^i_j \le K^i_j
\end{equation}
We also assume non-negativity for all counts: $n^i_j \ge 0$, $I_j \ge 0$, and $x^l_k \ge 0$. Finally, we assume that all cars travel at a constant speed, $v$.

The number of cars departing from an outgoing lane, $x^i_j$, depends on the junction index $j$, the current phase, and the time $T_g$ for which that phase is active. If $\mu^i_j$ is the saturation flow rate (service rate) for lane $i$ at junction $j$, the number of departing cars can be expressed as an integral of this flow over the green time, constrained by the queue length. A simplified model is:
\begin{equation}
    x^i_j = \int_{t}^{t+T_g} \mu^i_j \cdot \mathbf{1}[N^i_j(\tau) > 0] \cdot \mathbf{1}[\text{Phase}_i \text{ is green}] \,d\tau
    \label{eq:flow_out}
\end{equation}
where $\mathbf{1}[\cdot]$ is the indicator function, ensuring cars only flow when the lane is green and the queue is not empty.

\begin{figure}[htbp]
    \centering
    \includegraphics[width=0.7\textwidth]{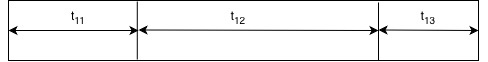}
    \caption{A single phase cycle for a junction $j$, showing the green time for each phase (e.g., $t^1_j, t^2_j, t^3_j$).}
    \label{fig:signal_timing}
\end{figure}

\vspace{1em} 
In the Q-learning algorithm, the Q-values $Q_t$ are updated each time a full phase cycle is completed. We will first analyze an incoming lane of junction 1 that is fed by an external source (e.g., lane 1). At time $t$, the queue is $N^1_1(t) = n^1_1$. At the end of the full cycle $T_c = t^1_1 + t^2_1 + t^3_1$, the queue at time $t+T_c$ is $N^1_1(t+T_c) = n^1_1 + I_1$, where $I_1$ is the number of arrivals.

Over this cycle, the expected queue length for this lane (which is green only during $t^1_1$) can be modeled. The expected number of arrivals during the red phases ($t^2_1 + t^3_1$) is $r_1(t^2_1 + t^3_1)$, and the expected number of departures during the green phase ($t^1_1$) is $v t^1_1$ (assuming a constant service rate $v$ and a non-empty queue). Thus, the change in expected queue length is:
\begin{equation}
    E[\Delta N^1_1] = E[N^1_1(t+T_c)] - N^1_1(t) \approx r_1(t^2_1 + t^3_1) - v t^1_1
\end{equation}
Here, we assume the expected value remains non-negative. If $r_1(t^2_1 + t^3_1) - v t^1_1 > 0$, congestion will tend to increase, and if it is negative, congestion will tend to decrease.

Now, let us consider an incoming lane at junction 1 that is fed by another junction (e.g., lane 2, $N^2_1$). Its queue evolution depends on the outflow from neighboring junctions (e.g., from junction 2, lanes 1 and 2, during their green phases $t^1_2$ and $t^2_2$) and its own outflow (during its green phase $t^2_1$). The expected queue length at the end of the cycle can be approximated as:
\begin{equation}
\begin{split}
    E[N^2_1(t+T_c)] \approx n^2_1 & + \alpha_{2,1} P(N^1_2 > 0) v t^1_2 \\
                               & + \alpha_{2,2} P(N^2_2 > 0) v t^2_2 - v t^2_1
\end{split}
\label{eq:internal_exp_queue}
\end{equation}
where $n^2_1 = N^2_1(t)$, $\alpha_{k,i}$ is the proportion of traffic from lane $i$ of junction $k$ that turns into lane 2 of junction 1, and the term $P(N^i_k > 0) v t^i_k$ approximates the expected number of cars arriving from the upstream junction's green phase.

A similar expression occurs for all other lane types. Here, we can see that the probability of a car from an incoming lane $i$ turning to an outgoing lane $k$ at junction $j$ is given by $\alpha^j_{ik}$. Thus, the sum of probabilities for a car in lane $i$ at junction $j$ to move to *any* of the possible outgoing lanes $k$ must be 1. Effectively, this means:
\begin{equation}
    \sum_k \alpha^j_{ik} = 1
\end{equation}
where the sum is over all valid outgoing lanes $k$ for an incoming lane $i$.

Now, in any one of the junctions (say, junction 1), we can evaluate the expected cost. Assuming a three-junction neighborhood where the cost is the sum of all queue lengths, $N_j = \sum_i N^i_j$, we have:
\begin{equation}
    E[c_1(t)] = \frac{1}{|N_1|} E\left[ \sum_{j \in N_1} \sum_{i=1}^{L_j} N^i_j \right]
\end{equation}
Assuming $|N_1| = 3$ (all junctions) and that the cost is the same for all junctions, we find:
\begin{equation}
    E[c_1(t)] = \frac{1}{3} E\left[ \sum_{j=1}^{3} \sum_{i=1}^{L_j} N^i_j \right]
\end{equation}
This can be further expanded by substituting the expected change for each lane, similar to the equations above. If we let $\bar{n} = \sum_{j=1}^{3} \sum_{i=1}^{L_j} n^i_j$ be the total base queue length, the expected cost becomes:
\begin{equation}
\begin{split}
    E[c_1(t)] = \frac{1}{3} \Big[ \bar{n} & + (r_1 - v)t^1_1 + \alpha^{2}_{13} P(N^1_2 > 0)vt^1_2 + \alpha^{2}_{23} P(N^2_2 > 0)vt^2_2 \\
    & + (r_1 - v)t^2_1 + \alpha^{3}_{12} P(N^1_3 > 0)vt^1_3 + \alpha^{3}_{32} P(N^3_3 > 0)vt^3_3 \\
    & + (r_1 - v)t^3_1 + (r_2 - v)t^1_2 + \alpha^{3}_{13} P(N^1_3 > 0)vt^1_3 + \alpha^{3}_{23} P(N^2_3 > 0)vt^2_3 \\
    & + (r_2 - v)t^2_2 + \alpha^{1}_{32} P(N^3_1 > 0)vt^3_1 + \alpha^{1}_{12} P(N^1_1 > 0)vt^1_1 \\
    & + (r_2 - v)t^3_2 + (r_3 - v)t^1_3 + \alpha^{1}_{13} P(N^1_1 > 0)vt^1_1 + \alpha^{1}_{23} P(N^2_1 > 0)vt^2_1 \\
    & + (r_3 - v)t^2_3 + \alpha^{2}_{12} P(N^1_2 > 0)vt^1_2 + \alpha^{2}_{32} P(N^3_2 > 0)vt^3_2 \\
    & + (r_3 - v)t^3_3 \Big]
\end{split}
\label{eq:full_cost}
\end{equation}

\vspace{1em}
At junction 1, when phase 1 is active, we can analyze the change in cost. Let $\delta c(t)$ be the change in the expected cost $E[c(t)]$ during one cycle. Assuming all other terms in the cost equation sum to a constant $K$, the change attributable to junction 1's phase 1 is:
\begin{equation}
    \delta c(t) \approx \frac{1}{3} \left[ K + (r_1 - v)t^1_1 + P(N^1_1 > 0)v t^1_1 \right]
\end{equation}
We do not know the sign of $K$, but we can control the other two terms by adjusting $t^1_1$. We can analyze the sign of the coefficient of $t^1_1$:
\begin{equation}
    (r_1 - v + \alpha^1_{13} P(N^1_1 > 0)v)
\end{equation}

If $(r_1 - v + P(N^1_1 > 0)v) < 0$, then the cost is minimized by *increasing* $t^1_1$. Conversely, if this term is positive, $t^1_1$ should be decreased to minimize the cost.

Thus, each individual state-action pair $(s_j, a_j)$ (which corresponds to setting the time $t^i_j$) should be chosen such that this expected cost is minimized. Only then can we say the total cost will be minimized.

Now, each agent acts observing only its current local state. In this particular problem, we notice that all three agents share the exact same cost function $c(t)$. Each agent, however, controls a different set of terms (its own phase durations $t^i_j$) in this common cost function. The Q-value for an agent is minimized if the expected cost is minimized. Therefore, the multiple agents are implicitly making a joint action to reduce the cost among their neighbors.

Assuming a specific state where junction 1 has phase 1 active, junction 2 has phase 2 active, and junction 3 has phase 3 active, the change in cost $\delta c_1(t)$ is a joint minimization problem. It becomes:
\begin{equation}
\begin{split}
    \delta c_1(t) \approx \frac{1}{3} \Big[ K & + (r_1 - v)t^1_1 + P(N^1_1 > 0)v t^1_1 \\
    & + (r_2 - v)t^2_2 + \alpha^2_{23} P(N^2_2 > 0)v t^2_2 \\
    & + (r_3 - v)t^3_3 + \alpha^3_{23} P(N^3_3 > 0)v t^3_3 \Big]
\end{split}
\end{equation}
Thus, the problem is a joint minimization over the variables $t^1_1$, $t^2_2$, and $t^3_3$.

\section{General Stochastic Model}
The 3-junction network in Section~\ref{example3junction} provides a concrete example of the underlying stochastic process. We can now generalize this model to a network of $J$ junctions, each with an arbitrary number of lanes and phases.

Let $j \in \{1, \ldots, J\}$ be the index for a junction.
Let $L_j$ be the number of incoming lanes at junction $j$.
Let $P_j$ be the number of phases in the RR schedule for junction $j$.
Let $t^p_j$ be the duration of phase $p \in \{1, \ldots, P_j\}$ at junction $j$.
The total cycle time for junction $j$ is $T_{c,j} = \sum_{p=1}^{P_j} t^p_j$.

We can categorize the set of incoming lanes $L_j$ into two disjoint sets:
\begin{itemize}
    \item $L_{j,ext}$: The set of lanes fed by external sources (e.g., lane 1 in the 3-junction example).
    \item $L_{j,int}$: The set of lanes fed by other junctions in the network.
\end{itemize}

\subsection{Queue Evolution for External Lanes}
For any lane $i \in L_{j,ext}$, let its associated phase be $p(i)$. The lane receives arrivals at an average rate $r_i$ and is served at a rate $\mu_i$ (e.g., $v$) during its green time $t^{p(i)}_j$. The expected change in queue length over one full cycle $T_{c,j}$ can be approximated as:
\begin{equation}
    E[\Delta N^i_j] \approx r_i (T_{c,j} - t^{p(i)}_j) - \mu_i P(N^i_j > 0) t^{p(i)}_j
    \label{eq:gen_ext_queue}
\end{equation}
The term $r_i (T_{c,j} - t^{p(i)}_j)$ represents the expected arrivals during the lane's "red" time (i.e., all phases other than its own). The term $\mu_i P(N^i_j > 0) t^{p(i)}_j$ represents the expected departures during its green time.

\subsection{Queue Evolution for Internal Lanes}
For any lane $i \in L_{j,int}$, the arrivals depend on the departures from one or more upstream junctions. Let $U(i)$ be the set of all upstream (junction, lane) pairs $(k, l)$ that feed into lane $(j, i)$. The expected change in queue length is:
\begin{equation}
\begin{split}
    E[\Delta N^i_j] \approx \sum_{(k,l) \in U(i)} \Big( \alpha^k_{l \to i} \mu_l P(N^l_k > 0) t^{p(l)}_k \Big) \\
    - \mu_i P(N^i_j > 0) t^{p(i)}_j
\end{split}
\label{eq:gen_int_queue}
\end{equation}
Here, $\alpha^k_{l \to i}$ is the turning probability from lane $l$ at junction $k$ into lane $i$ at junction $j$. The summation term represents the total expected arrivals from all upstream lanes during their respective green phases $t^{p(l)}_k$. The second term is the expected departure from lane $i$ itself.

\subsection{Generalized Cost Function}
The expected cost for an agent $j$, which is the average queue length across its neighborhood $N_j$, can now be written in its general form by summing the expected queue lengths of all lanes in all neighboring junctions:
\begin{equation}
    E[c_j(t)] = \frac{1}{|N_j|} \sum_{k \in N_j} \sum_{i=1}^{L_k} E[N^i_k(t)]
\end{equation}
The full expansion of this cost, similar to \eqref{eq:full_cost}, would involve substituting the queue evolution equations \eqref{eq:gen_ext_queue} and \eqref{eq:gen_int_queue} for every lane in the neighborhood. This confirms that the cost for agent $j$ is a complex function of its own action (phase durations $\{t^p_j\}$) and the actions of all other agents in the network $\{t^p_k\}_{k \ne j}$ that influence its neighborhood. This explicitly frames the problem as a multi-agent coordination game where each agent $j$ seeks to find the optimal durations $\{t^p_j\}_{p=1}^{P_j}$ that minimize this shared, coupled cost function.

\section{Stochastic Approximation Setting}
\subsection{Basic Stochastic Approximation Setting}
The basic stochastic approximation setting starts with an iterative scheme in $\mathbb{R}^d$ given by:
\begin{equation}
    x_{n+1} = x_n + a(n)[h(x_n) + M_{n+1}], \quad n \ge 0,
\end{equation}
where $h(\cdot)$ is a function whose fixed point we are trying to find by searching over $x_n$. $M_{n+1}$ is the noise obtained in addition to our function. Generally, we only observe the noisy iterates, $h(x_n) + M_{n+1}$, for our stochastic approximation setting. The following four assumptions are standard for the convergence of this scheme:

\begin{description}
    \item[(A1)] The map $h: \mathbb{R}^d \to \mathbb{R}^d$ is Lipschitz, i.e.,
    \begin{equation}
        |h(x) - h(y)| \le L|x - y| \quad \text{for some } 0 < L < \infty 
        \label{lipscitz1}
    \end{equation}

    \item[(A2)] The step sizes $\{a(n)\}$ are positive scalars satisfying:
    \begin{equation}
        \sum_n a(n) = \infty, \quad \sum_n a(n)^2 < \infty 
        \label{stepsize1}
    \end{equation}

    \item[(A3)] $\{M_n\}$ is a martingale difference sequence with respect to the increasing family of $\sigma$-fields $\mathcal{F}_n = \sigma(x_m, M_m, m \le n) = \sigma(x_0, M_1, \ldots, M_n), n \ge 0$, i.e.,
    \begin{equation}
        E[M_{n+1} | \mathcal{F}_n] = 0, \quad \text{a.s.}, n \ge 0 
        \label{martingalemeanzero}
    \end{equation}
    
    and $\{M_n\}$ is square integrable, satisfying:
    \begin{equation}
         E[|M_{n+1}|^2 | \mathcal{F}_n] \le K(1 + |x_n|^2), \quad \text{a.s.}, n \ge 0 
         \label{Martingalesquarege}
    \end{equation}
    
    for some constant $K > 0$.
    
    \item[(A4)] The iterates remain bounded, i.e.,
    \begin{equation}
        \sup_n |x_n| < \infty, \quad \text{a.s.}
        \label{iteratesbounded}
    \end{equation}
\end{description}

All of these four assumptions are very important. It can be shown that when all these assumptions are satisfied, the iterates will asymptotically track the ODE \cite{Borkar2008StochasticAA}.
\begin{equation}
    \dot{x}(t) = h(x(t)), \quad t \ge 0
    \label{eq:ode}
\end{equation}

\subsection{Stochastic Approximation in the Multi-Agent Setting}

There are many models of stochastic approximation in the multi-agent setting. We will focus on one for now, based on \cite{Borkar2008StochasticAA}. It is given by the ODE:
\begin{equation}
    \dot{x}(t) = h(x(t))
\end{equation}
with the condition:
\begin{equation}
    \frac{\partial h_i}{\partial x_j} > 0, \quad \forall j \ne i
\end{equation}

These are called "co-operative" ODEs, with the idea being that an increase in the $i$-th component (corresponding to some desirable quantity for the $i$-th agent) will lead to an increase in the $j$-th component as well, for $j \ne i$. Suppose the trajectories remain bounded. Then, a theorem by Hirsch (see \cite{HirschMorrisConvergence}) states that for almost all initial conditions, $x(t)$ converges to an equilibrium. That is, cooperative dynamics with bounded trajectories generically guarantee convergence rather than oscillation or chaos.

\subsection{Stochastic Approximation in the Traffic Control Setting}

To analyze the multi-agent traffic system, we adopt the stochastic approximation framework, drawing analogies to the Borkar model. We define the comprehensive state vector, $\bar{Q}$, as the collection of all $Q$-values for every state-action pair across all junctions in the network. The evolution of the system's mean-field dynamics can be described by the following Ordinary Differential Equation (ODE):
\begin{equation}
    \dot{\bar{Q}}(t) = \bar{h}(\bar{Q}(t))
    \label{eq:meanddriftQODE}
\end{equation}

where $\bar{h}$ represents the mean drift field.

The algorithm's dynamics are driven by the individual Q-learning update rule for each agent $j$. As established earlier, this update is given by:
\begin{equation}
\begin{split}
    Q_{j, t+1}(s_j, a_j) &= Q_{j,t}(s_j, a_j) + \gamma(t) \left( c_j(t) + \beta \min_{b \in \mathcal{A}} Q_{j,t}(s'_j, b) - Q_{j,t}(s_j, a_j) \right) \\
    &= (1 - \gamma(t)) Q_{j,t}(s_j, a_j) + \gamma(t) \left( c_j(t) + \beta \min_{b \in \mathcal{A}} Q_{j,t}(s'_j, b) \right)
\end{split}
\label{eq:q_update_split}
\end{equation}
This form illustrates that the new $Q$-value is a convex combination of the previous value and the current target value.

To frame this as a standard stochastic approximation, we must decompose the update into its deterministic drift and stochastic noise components. Let $F_{j,t}(\bar{Q})$ represent the stochastic update term:
\begin{equation}
    F_{j,t}(\bar{Q}) = c_j(t) + \beta \min_{b \in \mathcal{A}} Q_{j,t}(s'_j, b) - Q_{j,t}(s_j, a_j)
    \label{eq:stochasticupdatedrift}
\end{equation}

We define the mean drift $h_j(\bar{Q})$ as the conditional expectation of this term, given the history (filtration) $\mathcal{F}_t$, which includes the current state $\bar{Q}_t$:
\begin{equation}
    h_j(\bar{Q}_t) = \mathbb{E}\left[ F_{j,t}(\bar{Q}) \mid \mathcal{F}_t \right]
    \label{stochasticupdate1}
\end{equation}

By adding and subtracting this mean drift term inside the update rule \eqref{eq:q_update_split}, we get:
\begin{equation}
\begin{split}
    Q_{j, t+1}(s_j, a_j) = Q_{j,t}(s_j, a_j) &+ \gamma(t) h_j(\bar{Q}_t) \\
    &+ \gamma(t) \left( F_{j,t}(\bar{Q}) - h_j(\bar{Q}_t) \right)
\end{split}
\label{stochasticupdate2}
\end{equation}
This formulation clearly separates the update into two parts:
\begin{enumerate}
    \item \textbf{Drift Term:} $h_j(\bar{Q}_t)$, which corresponds to the vector field $\bar{h}$ in the ODE.
    \item \textbf{Noise Term:} $M_{t+1} = F_{j,t}(\bar{Q}) - h_j(\bar{Q}_t)$. This term is a martingale difference sequence by construction, as $\mathbb{E}[M_{t+1} \mid \mathcal{F}_t] = 0$.
\end{enumerate}

Based on the convergence theorems for stochastic approximation (\cite{Kushner1978}), the algorithm $\bar{Q}_{t}$ is guaranteed to converge to the set of stable fixed points of the ODE ($\dot{\bar{Q}} = \bar{h}(\bar{Q})$), provided the standard conditions hold: (i) the step sizes satisfy $\sum_{t=0}^{\infty} \gamma(t) = \infty$ and $\sum_{t=0}^{\infty} \gamma(t)^2 < \infty$, (ii) the noise $M_{t+1} = F_{j,t}(\bar{Q}) - h_j(\bar{Q}_t)$ is a martingale difference sequence (as shown above), (iii) the iterates $\{\bar{Q}_t\}$ remain bounded almost surely, and (iv) the mean drift $\bar{h}(\cdot)$ is Lipschitz continuous.

\section{Convergence Proof}
\subsection{Value Iteration}
The basic value iteration method as a vector method is given by:
\begin{equation}
    F\bar{J} = \bar{c} + \beta P\bar{J}
    \label{eq:value_iteration}
\end{equation}
where
\begin{equation}
    P = \begin{bmatrix}
    p_{11} & p_{12} & \cdots \\
    p_{21} & p_{22} & \cdots \\
    \vdots & \vdots & \ddots
    \end{bmatrix}
    \label{eq:p_matrix}
\end{equation}
\begin{equation}
    \bar{c} = \begin{bmatrix}
    \sum_{j \in S} P_{1j} c_{1j}(t) \\
    \sum_{j \in S} P_{2j} c_{2j}(t) \\
    \vdots
    \end{bmatrix}
    \label{eq:c_vector}
\end{equation}

\subsection{Asynchronous Value Iteration}
The Asynchronous version of the value iteration method says that
\begin{equation}
    J_{k+1}(i) = \begin{cases}
    (TJ_k)(i), & \text{if } i = i_k \\
    J_k(i), & \text{otherwise}
    \end{cases}
    \label{eq:async_update}
\end{equation}
where for $\beta < 1$
\begin{equation}
    (TJ)(i) = \min_{u \in \mathcal{U}(i)} \sum_{j=0}^{n} p_{ij}(u) (g(i,u,j) + \beta J(j))
    \label{eq:bellman_operator}
\end{equation}
It can be proven that this method will converge as long as all states are visited infinite times  \cite{books/lib/BertsekasT96} .

\subsection{Stochastic Approximation Modification}
A slightly different stochastic approximation method:
\begin{equation}
x_i(t+1) = x_i(t) + \gamma_i(t)\left(F_i(x(t)) - x_i(t) + w_i(t)\right)
\label{eq:stoch_approx_mod}
\end{equation}

where $x = \{x_1, x_2, \dots, x_n\} \in \mathbb{R}^n$ and $F = \{F_1, F_2, \dots, F_n\}$ are mappings from $\mathbb{R}^n$ to $\mathbb{R}$ and $w_i$ is a small random noise term.
This algorithm is also seen to converge \cite{Tsitsiklis1994}.

A slight modification to the Lipschitz assumption is given by
\begin{equation}
    \|F(x) - x^*\|_v \le \beta \|x - x^*\|_v
    \label{eq:lipschitz_mod}
\end{equation}

\subsection{Q-Learning Convergence Proof}

For value iteration, the T operator is given by

\begin{equation}
T_i(V) = \min_{u \in U(i)} E[c_{iu}] + \beta \sum_{j \in S} \rho_{ij}(u) V_j
\label{eq:t_operator}
\end{equation}

The Q-learning method is based on a modification of the Bellman equation $V^* = T(V^*)$.

Let $P = \{(i,u) \mid i \in S, u \in U(i)\}$ be the set of all state-action pairs and let $n$ be its cardinality.

Let after $t$ iterations, the vector $Q(t) \in \mathbb{R}^n$, with components $Q_{iu}(t), (i,u) \in P$ be updated according to the formula:

\begin{equation}
Q_{iu}(t+1) = Q_{iu}(t) + \gamma_{iu}(t) \left[ c_{iu} + \beta \min_{v \in U(s(i,u))} Q_{s(i,u),v}(t) - Q_{iu}(t) \right]
\label{eq:q_update}
\end{equation}

We now argue that this equation has the form of \eqref{eq:stoch_approx_mod}. Let $F$ be the mapping defined from $\mathbb{R}^n$ onto itself with components $F_{iu}$ defined by

\begin{equation}
F_{iu}(Q) = E[c_{iu}] + \beta E\left[\min_{v \in U(s(i,u))} Q_{s(i,u),v}\right]
\label{eq:f_operator}
\end{equation}

and

\begin{equation}
E\left[\min_{v \in U(s(i,u))} Q_{s(i,u),v}\right] = \sum_{j \in S} p_{ij}(u) \min_{v \in U(j)} Q_{jv}
\label{eq:f_operator_expectation}
\end{equation}

In view of \eqref{eq:f_operator_expectation}, \eqref{eq:f_operator}, the Q-learning update can be written as

\begin{equation}
Q_{iu}(t+1) = Q_{iu}(t) + \gamma_{iu}(t) \left( F_{iu}(Q(t)) - Q_{iu}(t) + w_{iu}(t) \right)
\label{eq:q_update_stochastic}
\end{equation}

where

\begin{equation}
\begin{split}
w_{iu}(t) = c_{iu} - E[c_{iu}] &+ \min_{v \in U(s(i,u))} Q(s(i,u),v)(t) \\
&- E\left[\min_{v \in U(s(i,u))} Q(s(i,u),v)(t) \mid \mathcal{F}(t)\right]
\end{split}
\label{eq:noise_term}
\end{equation}

The expectation in the expression $E(\min_{v \in U(s(i,u))} Q_{(s(i,u),v)}(t) \mid \mathcal{F}(t))$ is with respect to $s(i,u)$.

The vector form of $F(\bar{Q})$, where $n_s$ is the number of states and $n_a$ is the number of actions, can be written as:

\begin{equation}
\begin{bmatrix}
F_{1,a_1}(Q) \\
F_{1,a_2}(Q) \\
\vdots \\
F_{n_s, a_{n_a}}(Q)
\end{bmatrix}
=
\begin{bmatrix}
E[c_{1,a_1}] \\
E[c_{1,a_2}] \\
\vdots \\
E[c_{n_s, a_{n_a}}]
\end{bmatrix}
+ \beta
\begin{bmatrix}
\sum_{j \in S} p_{1j}(a_1) \min_{v \in U(j)} Q_{jv} \\
\sum_{j \in S} p_{1j}(a_2) \min_{v \in U(j)} Q_{jv} \\
\vdots \\
\sum_{j \in S} p_{n_s,j}(a_{n_a}) \min_{v \in U(j)} Q_{jv}
\end{bmatrix}
\label{eq:vector_form_f}
\end{equation}

which in matrix notation becomes $\bar{F}(\bar{Q}) = E[\bar{c}] + \beta P \bar{m}$, where

\begin{equation}
P = \begin{bmatrix}
P_{11}(a_1) & P_{12}(a_1) & \cdots & P_{1n_s}(a_1) \\
P_{11}(a_2) & P_{12}(a_2) & \cdots & P_{1n_s}(a_2) \\
\vdots & \vdots & \ddots & \vdots
\end{bmatrix}
\label{eq:p_matrix_detailed}
\end{equation}

and $\bar{m} = [\min_{v \in U(1)} Q_{1v}, \min_{v \in U(2)} Q_{2v}, \ldots, \min_{v \in U(n_s)} Q_{n_s,v}]^T$.

Following \cite{Tsitsiklis1994}, taking conditional variance on both sides of \eqref{eq:noise_term}, we find that
\begin{equation}
    E\left[ \|w_{iu}(t)\|^2 \mid \mathcal{F}(t) \right] \le \text{Var}(c_{iu}) + \max_{j \in S} \max_{v \in U(j)} Q_{jv}^2(t)
    \label{eq:variance_bound}
\end{equation}
For discounted problems ($\beta < 1$), \eqref{eq:f_operator} yields \cite{Tsitsiklis1994}:

\begin{equation}
    |F_{iu}(Q) - F_{iu}(Q')| \le \beta \max_{j \in S, v \in U(j)} |Q_{jv} - Q'_{jv}|, \quad \forall Q, Q'
    \label{eq:f_contraction}
\end{equation}

\subsection{Multi-agent Q-Learning Proof}
\begin{itemize}
    \item First we frame the multi-agent Q-learning problem as a vector update for a single state for the entire system.
    \item Then, we try to frame the problem as a large vector update over all states.
    \item Finally, we will state the problem as an asynchronous update of the large vector and show that it also satisfies our criteria for convergence.
\end{itemize}

\subsubsection{Single State update for the system}

We can think of our system as a network of nodes connected to each other. 
Thus, our system can be represented as a graph $G = (V, E)$, where $V$ is 
the set of $N$ junctions and $E$ is the set of roads connecting them. Let 
$\tilde{A} \in \mathbb{R}^{N \times N}$ denote the adjacency matrix with 
self-loops, where $\tilde{A}_{ij} = 1$ if junctions $i$ and $j$ are neighbors 
(i.e., share a road) or if $i = j$, and $\tilde{A}_{ij} = 0$ otherwise. Let 
$\tilde{D} = \text{diag}(|N_1|, \ldots, |N_N|)$ be the degree matrix, where 
$|N_j| = \sum_k \tilde{A}_{jk}$ is the number of junctions in the neighborhood 
of $j$ (including $j$ itself). The matrix $\tilde{D}^{-1}\tilde{A}$ then computes 
the average over each agent's neighborhood, consistent with the cost 
definition in \eqref{eq:cost}.

Define the vector of raw total queue lengths at each junction as:
\begin{equation}
\bar{\ell}(t) = \begin{bmatrix}
\sum_{i=1}^{L_1} q_i^1(t+1) \\
\sum_{i=1}^{L_2} q_i^2(t+1) \\
\vdots \\
\sum_{i=1}^{L_N} q_i^N(t+1)
\end{bmatrix}
\label{eq:raw_queue_vector}
\end{equation}

The cost vector for all agents can then be written as:
\begin{equation}
\bar{c}(t) = \tilde{D}^{-1}\tilde{A} \, \bar{\ell}(t)
\label{eq:cost_vector_graph}
\end{equation}

which recovers $c_j(t) = \frac{1}{|N_j|} \sum_{k \in N_j} \sum_{i=1}^{L_k} q_i^k(t+1)$ for each component.

At a given time step, suppose each agent $j$ is in state $s^j$ and takes action $a^j$. The full Q-learning update for the entire system can be written in vector form as:
\begin{equation}
\begin{bmatrix}
Q_{1,t+1}(s^1, a^1) \\
Q_{2,t+1}(s^2, a^2) \\
\vdots \\
Q_{N,t+1}(s^N, a^N)
\end{bmatrix}
=
\begin{bmatrix}
Q_{1,t}(s^1, a^1) \\
Q_{2,t}(s^2, a^2) \\
\vdots \\
Q_{N,t}(s^N, a^N)
\end{bmatrix}
+ \gamma(t) \left(
\tilde{D}^{-1}\tilde{A} \, \bar{\ell}(t)
+ \beta
\begin{bmatrix}
\min_{b} Q_{1,t}(s^{1'}, b) \\
\min_{b} Q_{2,t}(s^{2'}, b) \\
\vdots \\
\min_{b} Q_{N,t}(s^{N'}, b)
\end{bmatrix}
-
\begin{bmatrix}
Q_{1,t}(s^1, a^1) \\
Q_{2,t}(s^2, a^2) \\
\vdots \\
Q_{N,t}(s^N, a^N)
\end{bmatrix}
\right)
\label{eq:system_update}
\end{equation}

In compact notation:
\begin{equation}
\bar{Q}_{t+1} = \bar{Q}_t + \gamma(t) \left( \tilde{D}^{-1}\tilde{A} \, \bar{\ell}(t) + \beta \, \bar{m}(t) - \bar{Q}_t \right)
\label{eq:system_update_compact}
\end{equation}

where $\bar{Q}_t = [Q_{1,t}(s^1, a^1), \ldots, Q_{N,t}(s^N, a^N)]^T$ and $\bar{m}(t) = [\min_b Q_{1,t}(s^{1'}, b), \ldots, \min_b Q_{N,t}(s^{N'}, b)]^T$.

\subsubsection{Update over all agents, states, actions}
We define an analogous set named $P1 = \{(i, s_i, a_i) \mid i \in J, s_i \in S_i, a_i \in A_i\}$ with cardinality $n1$. Thus, the large vector is of the form $Q(t) \in \mathbb{R}^{n1}$ where $Q_{i,s_i,a_i}(t)$ update is of the form \eqref{eq:equationQmultiagent}.
\begin{itemize}
    \item This can also be brought to the form of \eqref{eq:stoch_approx_mod} by adding and subtracting the expectation term.
    \item Thus, we can write,
\end{itemize}
\begin{equation}
    Q_{j,s_j,a_j}(t+1) = Q_{j,s_j,a_j}(t) + \gamma(t) \left( F_{j,s_j,a_j}(Q) - Q_{j,s_j,a_j}(t) + w_{j,s_j,a_j}(t) \right)
    \label{eq:q_update_multi}
\end{equation}
where,
\begin{equation}
    F_{j,s_j,a_j}(Q) = E[c_{j,s_j,a_j}] + \beta E\left[\min_{v \in U(s(j,s_j,a_j))} Q_{s(j,s_j,a_j),v}\right]
    \label{eq:f_operator_multi}
\end{equation}
and here,
\begin{equation}
    E\left[\min_{v \in U(s(j,s_j,a_j))} Q_{s(j,s_j,a_j),v}\right] = \sum_{s'_j \in S_j} P_{j,s_j,s'_j}(u_j) \min_{v \in U(s'_j)} Q_{j,s'_j,v}
    \label{eq:f_expectation_multi}
\end{equation}
Also, the term $w_{j,s_j,a_j}$ can be written as:
\begin{equation}
\begin{split}
    w_{j,s_j,a_j} = c_{j,s_j,a_j} &+ \beta \min_{v \in U(s(j,s_j,a_j))} Q_{s(j,s_j,a_j),v} \\
    &- \left( E[c_{j,s_j,a_j}] + \beta E\left[\min_{v \in U(s(j,s_j,a_j))} Q_{s(j,s_j,a_j),v} \mid \mathcal{F}(t)\right] \right)
\end{split}
\label{eq:noise_term_multi}
\end{equation}
where
\begin{equation}
    \mathcal{F}(t) = \{Q_{j,s_j,a_j}(0), c_{j,s_j,a_j}(t) \forall j \in J, \forall s_j \in S_j, \forall a_j \in A_j, \forall t \in T\}
    \label{eq:filtration_multi}
\end{equation}
\subsubsection{Vector update forms}

The component-wise update established in \eqref{eq:q_update_multi}--\eqref{eq:noise_term_multi} can also be expressed in an expanded matrix form:

\begin{equation}
\begin{bmatrix}
F(Q_{t+1}(1_1, a_1^1)) \\ F(Q_{t+1}(2_1, a_1^2)) \\ \vdots \\ F(Q_{t+1}(1_1, a_2^1)) \\ F(Q_{t+1}(2_1, a_2^2)) \\ \vdots
\end{bmatrix}
=
\begin{bmatrix}
E[c_{\text{agent1}}(t)] \\
E[c_{\text{agent2}}(t)] \\
\vdots
\end{bmatrix}
+ \beta
\begin{bmatrix}
E[\min_{v \in U(s(1_1, a_1^1))} Q_{s(1_1, a_1^1),v}(t)] \\
E[\min_{v \in U(s(2_1, a_1^2))} Q_{s(2_1, a_1^2),v}(t)] \\
\vdots
\end{bmatrix}
\label{eq:f_vector_form}
\end{equation}

The last Expectation term can be written as:

\begin{equation}
\begin{split}
\begin{bmatrix}
E[\min_{v \in U(s(1_1, a_1^1))} Q_{s(1_1, a_1^1),v}(t)] \\
E[\min_{v \in U(s(2_1, a_1^2))} Q_{s(2_1, a_1^2),v}(t)] \\
\vdots
\end{bmatrix}
&= \\
\begin{bmatrix}
P_{1,1_1}(a_1^1) & 0 & \cdots & P_{1_1, 1_2}(a_1^1) & 0 & \cdots \\
0 & P_{2_1, 2_1}(a_1^2) & \cdots & 0 & P_{2_1, 2_2}(a_1^2) & \cdots \\
\vdots & \vdots & \ddots & \vdots & \vdots & \ddots
\end{bmatrix}
&
\begin{bmatrix}
\min_{v \in U(s(1_1, a_1^1))} Q \\
\min_{v \in U(s(2_1, a_1^2))} Q \\
\vdots \\
\min_{v \in U(s(1_1, a_2^1))} Q \\
\vdots
\end{bmatrix}
\end{split}
\label{eq:expectation_matrix_form}
\end{equation}

\subsection{Vector Update Lipschitz}
Thus, the vector update equation can be written in the form of:
\begin{equation}
    \bar{Q}^{t+1} = \bar{Q}^t + \gamma(\bar{F}(\bar{Q}^t) - \bar{Q}^t + \bar{w}^t)
    \label{eq:vector_update_form}
\end{equation}
where,
\begin{equation}
    \bar{F}(\bar{Q}^t) = E(\bar{c}^t) + \beta \bar{P}(\bar{Q'}_{n_s}^t)
    \label{eq:f_bar_definition}
\end{equation}
Thus,
\begin{equation}
    \bar{F}(\bar{Q}) - \bar{F}(\bar{Q}') = \beta P (\bar{Q}_{n_s} - \bar{Q'}_{n_s})
    \label{eq:f_bar_difference}
\end{equation}
Taking $\infty$-norm on both sides, we get,
\begin{equation}
    \| \bar{F}(\bar{Q}) - \bar{F}(\bar{Q}') \|_{\infty} \le \beta \| \bar{Q} - \bar{Q}' \|_{\infty}
    \label{eq:f_bar_contraction_norm}
\end{equation}
since,
\begin{equation}
    \|P\|_{\infty} \le 1
    \label{eq:p_norm_bound}
\end{equation}
This shows that the $F$ vector operator is Lipschitz, and hence, the operation is a contraction since $\beta < 1$.
Now, for the Asynchronous update.

\subsection{Asynchronous Update}

Now, the update term will be over a certain component of the state vector 
with a certain action, and we will show that the contraction property still 
holds in the asynchronous setting.

For a certain agent $j$, at a state $s_j$, with action $a_j$, we bound the 
difference of the $F$ operator evaluated at two different Q-vectors 
$\bar{Q}$ and $\bar{Q}'$:

\begin{equation}
\begin{split}
& |\bar{F}(\bar{Q})_{j,s_j,a_j} - \bar{F}(\bar{Q}')_{j,s_j,a_j}| \\
&= \beta \left| E\left[\min_b Q_j(s_j', b) - \min_b Q_j'(s_j', b)\right] \right| \\
&\le \beta E\left[ \left| \min_b Q_j(s_j', b) - \min_b Q_j'(s_j', b) \right| \right] \\
&\le \beta E\left[ \max_{b \in U(s_j')} |Q_j(s_j', b) - Q_j'(s_j', b)| \right] \\
&\le \beta \max_{j \in J} \max_{s_j' \in S_j} \max_{b \in U(s_j')} |Q_j(s_j', b) - Q_j'(s_j', b)| \\
&= \beta \|\bar{Q} - \bar{Q}'\|_\infty
\end{split}
\label{eq:f_bar_component_lipschitz}
\end{equation}

where the third inequality uses the fact that for any vectors $u, v$:
$|\min_b u_b - \min_b v_b| \le \max_b |u_b - v_b|$.

This is similar to the form used in \eqref{eq:f_contraction}, confirming that the 
component-wise operator $\bar{F}_{j,s_j,a_j}$ is a contraction in the infinity norm 
with modulus $\beta < 1$.

Next, we verify the noise conditions. The noise term $w_{j,s_j,a_j}(t)$ defined 
in \eqref{eq:noise_term_multi} satisfies:
\begin{equation}
E\left[ w_{j,s_j,a_j}(t) \mid \mathcal{F}(t) \right] = 0
\label{eq:noise_zero_mean_multi}
\end{equation}

by construction, since it is the difference between a random variable and its 
conditional expectation.

For the second moment, taking conditional expectation on both sides of 
\eqref{eq:noise_term_multi}, we obtain:

\begin{equation}
E\left[ \|w_{j,s_j,a_j}(t)\|^2 \mid \mathcal{F}(t) \right] \le \text{Var}(c_{j,s_j,a_j}) + \max_{j \in J} \max_{s_j \in S_j} \max_{v \in U(s_j)} Q_{j,s_j,v}^2(t)
\label{eq:noise_variance_bound_multi}
\end{equation}

which satisfies the square integrability condition of the form 
$E[\|w\|^2 \mid \mathcal{F}(t)] \le K(1 + \|\bar{Q}_t\|^2)$ required 
by assumption (A3) in Section 4.1.

Since $\beta$ < 1, the mapping $\bar{F}$ is a contraction with a unique fixed point $\bar{Q}^*$. Following \cite{Tsitsiklis1994}, the contraction property together with the variance bound \eqref{eq:noise_variance_bound_multi} guarantees the iterates remain bounded almost surely (satisfying assumption (A4)). Combined with the assumption that all state-action pairs 
are visited infinitely often with step sizes satisfying 
\eqref{eq:step_size_conditions}, all conditions of the stochastic 
approximation convergence theorem are met. Hence, the multi-agent 
Q-learning algorithm converges to the unique fixed point $\bar{Q}^*$ almost 
surely.

\bibliographystyle{plain}
\bibliography{references}


\appendix


\end{document}